\documentclass[conference]{IEEEtran}
\IEEEoverridecommandlockouts
\usepackage{amsmath,amssymb,amsfonts}
\usepackage{algorithmic}
\usepackage{graphicx}
\usepackage{textcomp}
\usepackage{caption}
\usepackage{subfig}
\usepackage{amsfonts}
\usepackage{url}
\usepackage{color}
\usepackage{hyperref}
\usepackage{listings}
\usepackage[ruled,linesnumbered,noend]{algorithm2e}
\usepackage[sort,nocompress]{cite} 

\usepackage[table,xcdraw]{xcolor}
\usepackage[bottom]{footmisc}

\def\BibTeX{{\rm B\kern-.05em{\sc i\kern-.025em b}\kern-.08em
		T\kern-.1667em\lower.7ex\hbox{E}\kern-.125emX}}
\newtheorem{definition}{Definition}
\newtheorem{example}{Example}
\newcommand{\orcid}[1]{\href{https://orcid.org/#1}{\includegraphics[width=8pt]{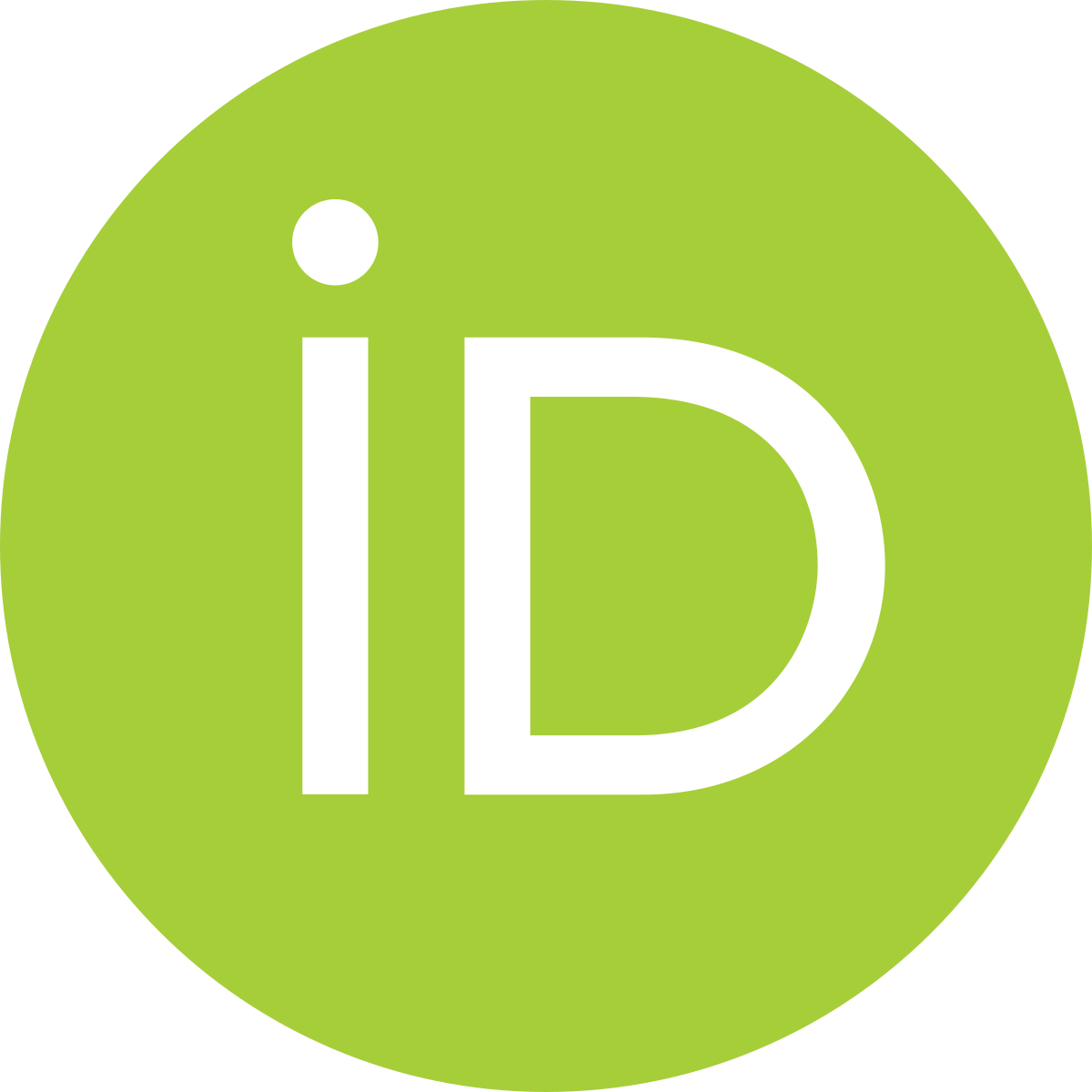}}}
\providecommand{\algorithmname}{Algorithmus}
\newcommand{\grayfont}{\color[HTML]{666666}}
\newcommand{\graycell}{\cellcolor[HTML]{C0C0C0}}

\begin{document}

\title{Object Type Clustering using Markov Directly-Follow Multigraph in Object-Centric Process Mining}

\author{\IEEEauthorblockN{Amin Jalali \orcid{0000-0002-6633-8587}}
	\IEEEauthorblockA{\textit{Department of Computer and Systems Sciences} \\
		\textit{Stockholm University}\\
		Stockholm, Sweden \\
		aj@dsv.su.se}
}	

\maketitle

\begin{abstract}
Object-centric process mining is a new paradigm with more realistic assumptions about underlying data by considering several case notions, e.g., an order handling process can be analyzed based on order, item, package, and route case notions. Including many case notions can result in a very complex model. To cope with such complexity, this paper introduces a new approach to cluster similar case notions based on Markov Directly-Follow Multigraph, which is an extended version of the well-known Directly-Follow Graph supported by many industrial and academic process mining tools. This graph is used to calculate a similarity matrix for discovering clusters of similar case notions based on a threshold. A threshold tuning algorithm is also defined to identify sets of different clusters that can be discovered based on different levels of similarity. Thus, the cluster discovery will not rely on merely analysts' assumptions. The approach is implemented and released as a part of a python library, called \textit{processmining}, and it is evaluated through a Purchase to Pay (P2P) object-centric event log file. 
Some discovered clusters are evaluated by discovering Directly Follow-Multigraph by flattening the log based on the clusters. 
The similarity between identified clusters is also evaluated by calculating the similarity between the behavior of the process models discovered for each case notion using inductive miner based on footprints conformance checking.
\end{abstract}

\begin{IEEEkeywords}
	process mining, clustering, Markov, OCPM, DFG, OCEL
\end{IEEEkeywords}

\section{Introduction}\label{Sec:Introduction}

Recent studies challenge the idea of applying process mining based on only one case notion~\cite{van2019object,berti2022event,van2020discovering}. 
For example, a simple order handling process can have many potential case notions like order, item, package, and route, which enable analyzing the business process from different perspectives. 
Indeed, it is more realistic to consider an event to be related to several case notions as several business entities might get affected by performing an activity in a business process.

Object-Centric Event Log (OCEL)~\cite{ghahfarokhi2021ocel} is the standard for relating one event to multiple objects representing different case notions.
Object-Centric Process Mining is a new paradigm in process mining supporting several case notions when analyzing such log files. 
These logs are considered to be closer to information systems' data in reality~\cite{van2020discovering}.
There are a few studies that introduce process model discovery techniques from such log files, e.g., Directly-Follows Multigraph~\cite{van2019object} and Object-Centric Petri nets~\cite{van2020discovering}.

Directly-Follows Multigraph (DFM)~\cite{van2019object} is a graph that shows the relationship between activities in a business process by incorporating several case notions. 
Relations in DFM show how the control in the process can move from one activity to another based on a case notion.
It can be considered as an equivalent graph like the well-known Directly-Follows Graph (DFG) but incorporates several case notions. 

\begin{figure*}[t!] 
	\begin{center}
		\includegraphics[width=1\textwidth]{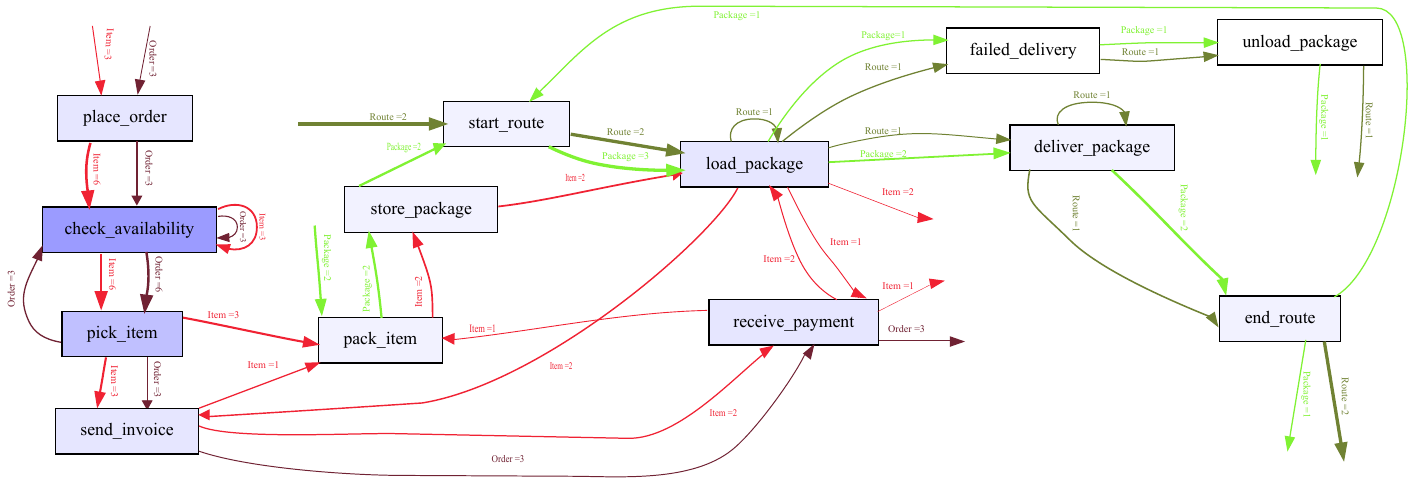}
		\caption{A Directly-Follows Multigraph (DFM), discovered from 39 events, indicates how process models incorporating all case notions can become complex.}
		\label{Fig:DFM}
	\end{center}
	\vspace{-1\baselineskip}
\end{figure*}

\figurename~\ref{Fig:DFM} shows an example of a DFM discovered from a toy example log file containing 39 events related to four case notions, i.e., item, order, package, and route, where their corresponding flows are colored by red, dark-red, green, and dark-green, respectively. 
The model is discovered using PM4Py~\cite{pm4py2019}, which is a python library that supports process mining.

As it can be seen in \figurename~\ref{Fig:DFM}, a DFM can easily become complex due to the existence of several case notions for each the process might have different underlying behavior. 
Separation of concerns is an approach to dealing with complexity in information systems~\cite{jalali2018hybrid,la2011managing}, which can be applied in this context by separating and classifying similar case notions into one category.
Discovering process models with several case notions that share similar behavior can simplify the models and enable analyzing interesting aspects from OCEL.

Therefore, this paper introduces a new approach to cluster similar case notions based on Markov Directly-Follow Multigraph. This graph is used to calculate a similarity matrix which enables clustering of the case notions based on a threshold. A threshold tuning algorithm is also defined to identify sets of different clusters that can be discovered based on different thresholds. The approach is implemented as a python library, and it is evaluated through a Purchase to Pay (P2P) object-centric event log file. 
Some discovered clusters are evaluated by discovering Directly Follow-Multigraph by flattening the log based on the clusters. 
The similarity between identified clusters is also evaluated by calculating the similarity between the behavior of the process models discovered for each case notion using inductive miner based on footprints conformance checking.

The rest of the paper is organized as follows.
Section~\ref{Sec:Background} gives a short background.
Section~\ref{Sec:Approach} formalizes the approach. 
Section~\ref{Sec:Implementation} elaborates on the implementation.
Section~\ref{Sec:Evaluation} reports the evaluation results.
Section~\ref{Sec:Conclusion} concludes the paper and introduces future research.

\section{Background}\label{Sec:Background}

This section summarizes the concepts needed to follow the rest of the paper. 

Process discovery is the most important use case of process mining~\cite{van2012process} that has received attention for many years.
The idea is to generate process models from event logs recording events during the enactment of a business process. 
Such logs require having a case identifier, activity name, and the order of events that happened (usually through a timestamp).
The case identifier represents the case notion based on which the behavior of process models can be identified.

Many commercial and open-source tools are available, which are developed under the assumption of having only one case notion in the log file.
Most of these tools focus on the generation of Directly-Follows Graphs (DFGs) as a means to visualize the control flow - which is used a lot by practitioners due to their simplicity~\cite{van2019practitioner}.
Although DFGs can be misleading due to lack of support for concurrency~\cite{van2019practitioner}, they can be helpful as an intermediate model to discover more advanced models as done by, e.g., Split Miner~\cite{augusto2019split}, Heuristics Miner~\cite{weijters2011flexible} and Fodina~\cite{vanden2017fodina}.
DFGs are also used in variant analysis where different models of a business process representing different variations can be compared to each other~\cite{taymouri2020business,jalali2021dfgcompare,taymouri2021business}.

In reality, a process can be analyzed using logs that contain several case notions, e.g., an order handling process can be analyzed based on order, item, package, and route case notions. 
Analysts used to \textit{flat} these logs to apply process mining techniques - built under the assumption of dealing with one case notion. 
Such flattening raises problems including \textit{convergence} and \textit{divergence}~\cite{van2019object}.

Transforming logs incorporating several case notions into one can cause problems.
An example of \textit{convergence} problem is repeating an event related to the occurrence of a batch job that handles many items - when flattening the log based on the item notion. 
It might enable discovering the batch activity in the discovered process model, but it can cause the problem of counting the wrong occurrence of the batch job activities. 
An example of a \textit{divergence} problem is losing the order between checking the availability of an item and picking it up when flattening the log based on the order notion. 
It can cause undesirable and incorrect loops as the order between the activities will be lost if removing the item notion, based on which the relation between checking the availability of an item and picking it up can be identified.

Object-Centric Event Log (OCEL)~\cite{ghahfarokhi2021ocel} is the standard that enables relating one event to multiple objects representing different case notions, and Object-Centric Process Mining (OCPM) is a new process mining paradigm that supports several case notions when analyzing such log files~\cite{van2019object}.
Directly Follow Multigraph (DFM) is one way to discover process models from OCEL, which is similar to DFG but supports different object types, representing different case notions~\cite{van2019object}.
Object-centric Petri nets is another discovery technique that can generate process models from OCEL~\cite{van2020discovering}.

From the tools support perspective, PM4Py~\cite{pm4py2019} is a python library that supports discovering DFM and object-centric Petri nets, and PM4Py-MDL is a python library that extends the functionality of PM4Py to support performance and conformance analysis through token-based replay~\cite{van2020discovering}. 
In addition, a stand-alone object-centric process cube tool is developed to support cube operations, i.e., slice and dice~\cite{ghahfarokhi2022python}. 
We also can see a rising interest in supporting OCPM by commercial tools, e.g., MEHRWERK Process Mining (MPM)~\cite{MPM2021}, which indicates how relevant is this problem in practice.

The tool support for OCPM is expanding not only in analysis but also in the pre-analysis phase, where data shall be Extracted, Transformed, and Loaded for conducting process mining. For example, a tool is developed to extract OCEL from ERP systems, i.e., SAP ERP System~\cite{berti2022event} which enables extracting OCELs from well-known processes in SAP ERP, e.g., Purchase to Pay (P2P) and Order to Cash (O2C).
Indeed, sample P2P and O2C logs in OCEL format are available through \url{http://ocel-standard.org}~\cite{ghahfarokhi2021ocel}, which empowers researchers to develop further artifacts and evaluate them based on these data.

The rise of big data introduces some challenges in applying process mining in practice, like scalability or discovering process models from logs that do not fit the memory of a computer~\cite{jalali2020graph}, which is also the case for OCPM.
Graph databases provide good capabilities to overcome this challenges~\cite {jalali2020graph,berti2022scalable}.
Several studies show how databases like Neo4j and MongoDB can be used to store and analyze both traditional and object-centric log files~\cite{jalali2020graph,berti2022scalable,esser2021multi}.

The application of OCPM techniques also requires adaptations in four competing quality dimensions of process mining, i.e., fitness, precision, simplicity, and generalization~\cite{adams2021precision}.
Adams J.N. and van der Aalst W.M.P. define how precision and fitness of object-centric Petri nets can be calculated by replaying the model with respect to an OCEL~\cite{adams2021precision}. 
Calculating these measures based on other techniques like alignment is still open for research, which is also the case for simplicity and generalization measures.

In summary, OCPM is a new paradigm that needs further research to be applied in practice. 
The current algorithms that enable discovering object-centric process models generate very complex process models. 
One way to deal with this complexity would be the separation of case notions into clusters based on their similarities. 
Such separation can also help future process discovery algorithms to consider object-type similarities when discovering process models from OCEL.
The next section explains how such separation can be performed using Directly Follow Multigraphs.

\section{Approach}\label{Sec:Approach}

\begin{figure}[b!] 
	\begin{center}
		\includegraphics[width=1\linewidth]{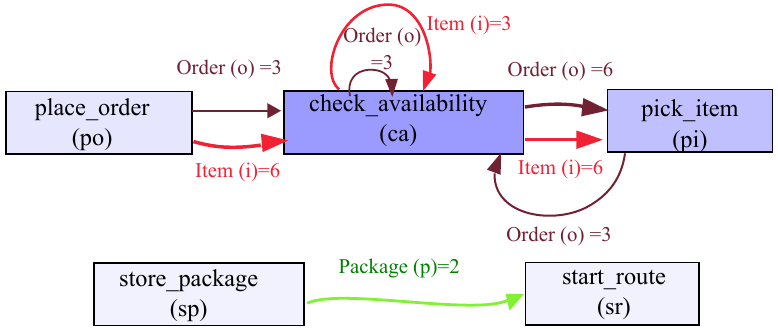}
		\caption{A simple DFM taken from \figurename~\ref{Fig:DFM} for explaining the approach.}
		\label{Fig:DFM_RunningExample}
	\end{center}
\end{figure}

This section defines the approach to identifying different clusters of similar case notions. 
To explain the definitions, a part of \figurename~\ref{Fig:DFM} will be used as a running example, shown in \figurename~\ref{Fig:DFM_RunningExample}.

For simplicity, acronyms are used instead of the activities' names, which are shown in parenthesis in the figure.
For example, we will use $po$ instead of $place\_order$, $o$ instead of $order$, and so on.

\begin{definition}[\textbf{Directly-Follows Multigraph (DFM)}]
	A Directly-Follows Multigraph (DFM) is a tuple $G=(OT, T, R, f)$, where:
	\begin{itemize}
		\item[-] $OT$ is the set of object types,
		\item[-] $T$ represents the set of tasks
		\item[-] $R=(T\times OT\times T)$ is the set of relations connecting two tasks based on an object type. We call the first task the source and the second one the target, representing the task from/to which the relation starts/ends, respectively.
		\item[-] $f\in R\rightarrow \mathbb{N}$ is a function that assigns a natural number, representing a frequency, to each relation.
	\end{itemize}
	Considering $\Theta\subseteq OT$ as a subset of object types, two operators on the graph's tasks can be defined as follow:
	\begin{itemize} 
		\item[-] $\overset{\Theta}\bullet t$ represents the operator that retrieves the set of tasks from which there are relations to task $t$ for an object types within $\Theta$, i.e.,:
		\subitem $\overset{\Theta}\bullet t=\{t^\prime\in T| \exists_{\theta\in \Theta}(t^\prime,\theta,t)\in R\}$.
		\item[-] $t\overset{\Theta}\bullet$  represents the operator that retrieves the set of tasks to   which there are relations from task $t$ for an object types within  $\Theta$, i.e.,:
		\subitem  $t\overset{\Theta}\bullet =\{t^\prime\in T| \exists_{\theta\in \Theta}(t,\theta,t^\prime)\in R\}$.
	\end{itemize}
\end{definition}

\begin{example}
	We can define the Directly-Follows Multigraph (DFM) for our running example in \figurename~\ref{Fig:DFM_RunningExample} as $G=(OT, T, R, f)$, where: 
	\begin{itemize}
		\item[-] $OT=\{o,i,p\}$ is the set of object types.
		\item[-] $T=\{po,ca,pi,sp,sr\}$ is the set of tasks.
		\item[-] $R=\{(po,o,ca), (po,i,ca), (ca,o,ca), (ca,i,ca), \\(ca,o,pi), (ca,i,pi), (pi,o,ca), (sp,p,sr)\}$ is the set of relations. $po$ is the source and $ca$ is the target of $(po,o,ca)$ relation.
		\item[-] $f((po,o,ca))=3$,  $f((po,i,ca))=6$, $f((ca,o,ca))=3$, $f((ca,i,ca))=3$, $f((ca,o,pi))=6$, $f((ca,i,pi))=6$, $f((pi,o,ca))=3$, $f((sp,p,sr))=2$ assigns frequencies to relations.
	\end{itemize}
	Examples of the operations based on the running example are given below:
	\begin{itemize}
		\item[-] $\overset{\{i\}}\bullet ca=\{po,ca\}$ retrieves a set of tasks from which there are outgoing flows to \textit{check availability} ($ca$) for object type \textit{item} ($i$). Note that we can have different result if we change the object type, i.e., $\overset{\{o\}}\bullet ca=\{po,ca,pi\}$ which  retrieves the set of tasks from which there is a relation to \textit{check availability} ($ca$) for object type \textit{order} ($o$). 
		\item[-] $ca\overset{\{i\}}\bullet=\{ca,pi\}$ and $po\overset{\{o\}}\bullet=\{ca\}$ retrieves a set of tasks to which there is a relation from \textit{check availability} ($ca$) using \textit{item} ($i$) object type and from \textit{place order} ($po$) using \textit{order} ($o$) object type, respectively.
	\end{itemize}
\end{example}

\begin{table*}[t!] 
	\begin{minipage}{.3\linewidth}
		\subfloat[Probability of relations for Item]
		{
			\resizebox{1\columnwidth}{!}{
				\begin{tabular}{c|c|c|c|c|c|}
					\cline{2-6}
					& ca  & pi  & po  & sp & sr \\ \hline
					\multicolumn{1}{|l|}{ca} & \textbf{1/3} & \textbf{2/3} & \grayfont{0} & \grayfont{0} & \grayfont{0} \\ \hline
					\multicolumn{1}{|l|}{pi} & \grayfont{0} & \grayfont{0} & \grayfont{0} & \grayfont{0} & \grayfont{0} \\ \hline
					\multicolumn{1}{|l|}{po} & \textbf{1} & \grayfont{0} & \grayfont{0} & \grayfont{0} & \grayfont{0} \\ \hline
					\multicolumn{1}{|l|}{sp} & \grayfont{0} & \grayfont{0} & \grayfont{0} & \grayfont{0} & \grayfont{0} \\ \hline
					\multicolumn{1}{|l|}{sr} & \grayfont{0} & \grayfont{0} & \grayfont{0} & \grayfont{0} & \grayfont{0}  \\ \hline
				\end{tabular}
		}}
	\end{minipage}%
	\hfill
	\begin{minipage}{.3\linewidth}
		\subfloat[Probability of relations for Order]
		{
			\resizebox{1\columnwidth}{!}{
				\begin{tabular}{c|c|c|c|c|c|}
					\cline{2-6}
					& ca  & pi  & po  & sp & sr \\ \hline
					\multicolumn{1}{|l|}{ca} & \textbf{1/3} & \textbf{2/3} & \grayfont{0} & \grayfont{0} & \grayfont{0} \\ \hline
					\multicolumn{1}{|l|}{pi} & \textbf{1} & \grayfont{0} & \grayfont{0} & \grayfont{0} & \grayfont{0} \\ \hline
					\multicolumn{1}{|l|}{po} & \textbf{1} & \grayfont{0} & \grayfont{0} & \grayfont{0} & \grayfont{0} \\ \hline
					\multicolumn{1}{|l|}{sp} & \grayfont{0} & \grayfont{0} & \grayfont{0} & \grayfont{0} & \grayfont{0}  \\ \hline
					\multicolumn{1}{|l|}{sr} & \grayfont{0} & \grayfont{0} & \grayfont{0} & \grayfont{0} & \grayfont{0}  \\ \hline
				\end{tabular}
		}}
	\end{minipage}%
	\hfill
	\begin{minipage}{.3\linewidth}
		\subfloat[Probability of relations for Package]
		{
			\resizebox{1\columnwidth}{!}{
				\begin{tabular}{c|c|c|c|c|c|}
					\cline{2-6}
					& ca  & pi  & po  & sp & sr \\ \hline
					\multicolumn{1}{|l|}{ca} & \grayfont{0} & \grayfont{0} & \grayfont{0} & \grayfont{0} & \grayfont{0} \\ \hline
					\multicolumn{1}{|l|}{pi} & \grayfont{0} & \grayfont{0} & \grayfont{0} & \grayfont{0} & \grayfont{0} \\ \hline
					\multicolumn{1}{|l|}{po} & \grayfont{0} & \grayfont{0} & \grayfont{0} & \grayfont{0} & \grayfont{0} \\ \hline
					\multicolumn{1}{|l|}{sp} & \grayfont{0} & \grayfont{0} & \grayfont{0} & \grayfont{0} & \textbf{1}  \\ \hline
					\multicolumn{1}{|l|}{sr} & \grayfont{0} & \grayfont{0} & \grayfont{0} & \grayfont{0} & \grayfont{0}  \\ \hline
				\end{tabular}
		}}
	\end{minipage}%
	\caption
	{%
		The probability of each relation is represented through a matrix per object type, where rows and columns represent the source and target task, respectively.%
		\label{Table:RunningExampleProbablities}%
	}%
\end{table*}

To find similarities between the control flow for different case notions, we convert the Directly Follow Multigraph to Markov Directly Follow Multigraph, defined below. Also, we define a similarity measure that calculates how similar the control flow of the process model is with respect to two given object types. 

\begin{definition}[\textbf{Markov Directly-Follows Multigraph (Markov DFM)}] 
	Let $G=(OT, T, R, f)$ be a DFM. $M=(G, p, sim)$ is a Markov DFM, where $p\in R\rightarrow [0$-$1]\subset\mathbb{Q}$ is a function that assigns a positive rational number between zero and one, representing the probability, to a relation. $sim \in OT\times OT\rightarrow [0$-$1]\subset\mathbb{Q}$ is a function that assigns a positive rational number between zero and one, representing the similarity, to an object types pair, where:
	\begin{equation}
		p\big((t,\theta,t^\prime)\big)\leftarrow 
		\frac{f\big((t,\theta,t^\prime)\big)}{ \sum_{\forall t^{''}\in t\overset{\{\theta\}}\bullet}{f\big((t,\theta,t^{''})\big)} }
	\end{equation}
	
	\begin{equation}
		\resizebox{1\hsize}{!}{$
			sim(\theta_1, \theta_2)\leftarrow 
			\cfrac{\sum_{\forall t,t^{'}\in T}{\big(p(t,\theta_1,t^{'})*p(t,\theta_2,t^{'})\big)}}{\sum_{\forall t_1,t_2\in T}{\big(\cfrac{p(t_1,\theta_1,t_2)^2+p(t_1,\theta_2,t_2)^2}{2}\big)}}
			$}
	\end{equation}
	
\end{definition}

We can define the Markov Directly-Follows Multigraph (DFM) for our running example as $M=\big(G=(OT, T, R, f), p, sim\big)$. Let's calculate  $p$ using an example.

\begin{example}
	\begin{itemize}
		\item[-] $p\big((ca,o,pi)\big)=f\big((ca,o,pi)\big) \Big/ \Big( \sum_{\forall t\in ca\overset{\{o\}}\bullet}{f\big((ca,o,t)\big)} \Big)$ 
		$=6\Big/ \Big( \sum_{\forall t\in {\{ca,pi\}}}{f\big((ca,o,t)\big)} \Big)$  
		$=6\Big/ \Big( f\big((ca,o,ca)\big)+f\big((ca,o,pi)\big) \Big) = 6\Big/ \Big(3+6\Big)=6/9=2/3$, which is the probablity of occurence of \textit{check availability} given \textit{place order} is occured for object type \textit{order} in this model. 
	\end{itemize}
\end{example}

\begin{figure}[t!] 
	\begin{center}
		\includegraphics[width=1\linewidth]{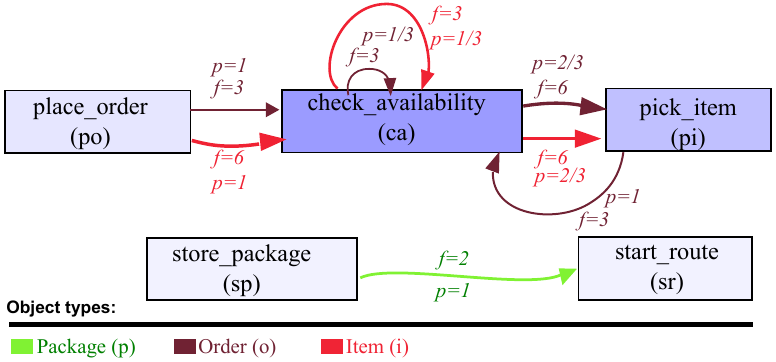}
		\caption{A Markov DFM of the DFM presented in \figurename~\ref{Fig:DFM_RunningExample}.}
		\label{Fig:DFM_RunningExampleProb}
	\end{center}
\end{figure}

We can illustrate our graph based on this definition visually through~\figurename~\ref{Fig:DFM_RunningExampleProb}, where the frequencies and probabilities of relations are shown by $p$ and $f$, respectively.
Note that probabilities can be represented by a matrix per object type, where rows and columns indicate the source and target tasks of a relation, as shown in \tablename~\ref{Table:RunningExampleProbablities}.
This table also makes it easier to explain the similarity calculation using $sim$ function.

\begin{example}
	As an example, let us to calculate $sim(i,o)$, where the probabilities of relations for item and order object types can be represented by $P_i$ and $P_o$ matrices as also shown in \tablename~\ref{Table:RunningExampleProbablities}.
	\begin{itemize}
		\item $
		P_i=\begin{bmatrix}
			\frac{1}{3} & \frac{2}{3} & 0 & 0 & 0 \\
			0 & 0 & 0 & 0 & 0 \\
			1 & 0 & 0 & 0 & 0 \\
			0 & 0 & 0 & 0 & 0 \\
			0 & 0 & 0 & 0 & 0 
		\end{bmatrix}
		,P_o=\begin{bmatrix}
			\frac{1}{3} & \frac{2}{3} & 0 & 0 & 0 \\
			1 & 0 & 0 & 0 & 0 \\
			1 & 0 & 0 & 0 & 0 \\
			0 & 0 & 0 & 0 & 0 \\
			0 & 0 & 0 & 0 & 0 
		\end{bmatrix}
		$
	\end{itemize}
	The similarity function calculates the similarity accordingly:
	\begin{itemize}
		\item[-] it calculates the denominator by summing up every element of $(P_i^2+P_o^2)/2$, which is equivalent to $		
		\sum_{\forall t_1,t_2\in T}{\big(\cfrac{p(t_1,i,t_2)^2+p(t_1,o,t_2)^2}{2}\big)}		
		$, which is eual to $\frac{37}{18}$.
		\item[-] The similarity will then be calculated by $P_i\cdot P_o $ devided by the calculated denominator, which will be equal to $\frac{252}{333}=0.76$.
	\end{itemize}
\end{example}

It is straightforward to calculate the similarity of \textit{Package} with \textit{Item} and also with \textit{Order} in our running example. As the numerator will always be zero, the similarity will be zero. The similarity result of the process for each object type pair for this example can be shown as a matrix, represented in \tablename~\ref{Table:Table1_SimilarityMatrix}.
We call this matrix the similarity matrix.

\begin{table}[h!] 
	\centering
	\begin{minipage}{0.5\linewidth}
		{
			\resizebox{1\columnwidth}{!}{				
				\begin{tabular}{c|c|c|c|}
					\cline{2-4}
					& o  & i & p   \\ \hline
					\multicolumn{1}{|l|}{o} & 1.0 & 0.76 & 0.0  \\ \hline
					\multicolumn{1}{|l|}{i} & 0.76 & 1.0 & 0.0  \\ \hline
					\multicolumn{1}{|l|}{p} & 0.0 & 0.0 & 1.0  \\ \hline
				\end{tabular}
		}}
	\end{minipage}
	\caption
	{
		Calculated Similarity Matrix that shows the similarity of the process for object type pairs.%
		\label{Table:Table1_SimilarityMatrix}%
	}
\end{table}

\begin{algorithm}[h!]
	\SetAlgoLined 
	\SetKwFunction{algo}{discoverClusters}
	
	\SetKwProg{myalg}{Algorithm}{}{}
	\myalg{\algo{$\big((OT, T, R, f), p, sim\big), threshold$}}{
		$clusters\leftarrow \{\}$\;
		\ForEach {$\theta_1,\theta_2\in OT$}{
			\If { $ sim(\theta_1, \theta_2) >= threshold$}{
				$X\leftarrow \bigcup_{C\in clusters}{\{C|\{\theta_1\}\subseteq C\  \vee \ \{\theta_2\}\subseteq C\}}$\;
				$clusters\leftarrow clusters\backslash X\  \cup\  \{\{\newline\bigcup_{C\in X}{C}\ \cup\ \{\theta_1,\theta_2\}\ \}\}$;
			}
			
		}
		\KwRet $clusters$\;{}
	}
	\caption{Clustering a Markov DFM algorithm}
	\label{Alg:Clustering}
\end{algorithm}

\algorithmname~\ref{Alg:Clustering} defines how clusters of similar object types can be discovered from a Markov DFM given a threshold. 
It defines an empty set for clusters of object types (line 2).
Then, for each pair of object types (line 3), if the similarity between them is greater or equal than the given threshold (line 4), it i) retrieves a union of sets of clusters that contains one of the object types (line 5), and ii) excludes the identified sets from the clusters and add all object types within these clusters in addition to two compared object types as a new cluster in the clusters set (line 6).
It finally returns the identified set of clusters (line 7).

\tablename~\ref{Table:SimilarityMatrixByThresholds} shows the result of calling this algorithm for the running example using different thresholds in addition to the filtered similarity matrix.

\begin{table}[h!] 
	\begin{minipage}{.33\linewidth}
		\subfloat[\centering 1 cluster when threshold=0, i.e., $\{\{i,o,p\}\}$]
		{
			\resizebox{1\columnwidth}{!}{				
				\begin{tabular}{c|c|c|c|}
					\cline{2-4}
					& o  & i & p   \\ \hline
					\multicolumn{1}{|l|}{o} & \graycell{1.0} & \graycell{0.76} & \graycell{0.0}  \\ \hline
					\multicolumn{1}{|l|}{i} & \graycell{0.76} & \graycell{1.0} & \graycell{0.0}  \\ \hline
					\multicolumn{1}{|l|}{p} & \graycell{0.0} & \graycell{0.0} & \graycell{1.0}  \\ \hline
				\end{tabular}
		}}
	\end{minipage}%
	\hfill
	\begin{minipage}{.33\linewidth}
		\subfloat[\centering 2 clusters when threshold=0.01, i.e., $\{\{i,o\},\{p\}\}$]
		{
			\resizebox{1\columnwidth}{!}{
				\begin{tabular}{c|c|c|c|}
					\cline{2-4}
					& o  & i & p   \\ \hline
					\multicolumn{1}{|l|}{o} & \graycell{1.0} & \graycell{0.76} &   \\ \hline
					\multicolumn{1}{|l|}{i} & \graycell{0.76} & \graycell{1.0} &   \\ \hline
					\multicolumn{1}{|l|}{p} &  &  & \graycell{1.0}  \\ \hline
				\end{tabular}
		}}
	\end{minipage}%
	\hfill
	\begin{minipage}{.33\linewidth}
		\subfloat[\centering 3 clusters when threshold=0.77, i.e., $\{\{i\},\{o\},\{p\}\}$]
		{
			\resizebox{1\columnwidth}{!}{
				\begin{tabular}{c|c|c|c|}
					\cline{2-4}
					& o  & i & p   \\ \hline
					\multicolumn{1}{|l|}{o} & \graycell{1.0} &  &   \\ \hline
					\multicolumn{1}{|l|}{i} &  & \graycell{1.0} &   \\ \hline
					\multicolumn{1}{|l|}{p} &  &  & \graycell{1.0}  \\ \hline
				\end{tabular}
		}}
	\end{minipage}%
	\caption
	{%
		Filtered similarity matrix and Identified clusters for the running example by setting different thresholds.%
		\label{Table:SimilarityMatrixByThresholds}%
	}%
\end{table}

\tablename~\ref{Table:SimilarityMatrixByThresholds}(a) represents the result and the filtered similarity matrix when we call the algorithm by setting the threshold to zero. In this case, we will receive only one cluster that includes all object types. It is because the value for $sim(\theta_1,\theta_2)$ is always greater or equal to zero (line 4 of the algorithm), so all object types will be added to the returned cluster, so the result for our running example will be $\{\{i,o,p\}\}$. 

\tablename~\ref{Table:SimilarityMatrixByThresholds}(b) represents the result and the filtered similarity matrix when we call the algorithm by setting the threshold to one percent. 
As seen in the filtered matrix, the relation between $p$ and other object types will be filtered as it is lower than the threshold. 
This result in the separation of these object types from others, so we will receive two clusters, i.e., $\{\{i,o\},\{p\}\}$. 

\tablename~\ref{Table:SimilarityMatrixByThresholds}(c) represents the result and the filtered similarity matrix when we call the algorithm by setting the threshold to 77 percent. 
This result is $\{\{i\},\{o\},\{p\}\}$. It can be said that if an object type in the filtered matrix has similarities to others, they will be in the same cluster.

\begin{algorithm}[h!]
	\SetAlgoLined 
	\SetKwFunction{algo}{tuneClusters}
	
	\SetKwProg{myalg}{Algorithm}{}{}
	\myalg{\algo{$M, threshold, res$}}{
		\If { $res=\{\}$}{
			$res\leftarrow \{(0,discoverClusters(M, 0))\}$\;
			$res\leftarrow res\ \cup\ \{(1,discoverClusters(M, 1))\}$\;
			\KwRet $tuneClusters(M, 0.5, res)$\;
		}
		\Else{
			\If { $(threshold,\_)\in res$}{
				\KwRet $res$\;
			}
			\Else{				
				$CT\leftarrow  discoverClusters(M, threshold)$\;
				$res\leftarrow res\ \cup\ \{(threshold,CT)\}$\;
				
				$u\leftarrow min\{i\ |\ \forall_{(i,-)\in res}\ i>threshold\}$\; 
				$l\leftarrow max\{i\ |\ \forall_{(i,-)\in res}\ i<threshold\}$\; 
				
				\If { $|\{C|\forall_{(t,C)\in res}\ t=u\}| \neq\ |CT|$}{ 
					$t\leftarrow round((threshold+u)/2,2)$\;
					$res\leftarrow res\ \cup\ \{(t,discoverClusters(M, t))\}$\;
				}

				\If { $|\{C|\forall_{(t,C)\in res}\ t=l\}| \neq\ |CT|$}{ 
					$t\leftarrow round((threshold+l)/2,2)$\;
					$res\leftarrow res\ \cup\ \{(t,discoverClusters(M, t))\}$\;
				}
				
				\KwRet $res$\;
			}
		}
	}
	\caption{Cluster tuning algorithm}
	\label{Alg:Tuning}
\end{algorithm}

In practice, it is difficult to change the threshold to find all possible clusters manually, so \algorithmname\ref{Alg:Tuning} tunes the threshold to identify all possibilities. 
This algorithm gets the $M$ (as a DFM), $threshold$, and $res$ - which is the result set representing the result of the previous tuning attempt. 
When calling this algorithm, the $res$ is an empty set as no tuning has happened. 
The algorithm performs recursively.

If it is the first time the algorithm is called, it discovers clusters for thresholds 0 and 1 and adds the result to the $res$.
Then, it calls itself to tune the cluster discovery based on 0.5 thresholds and calculated $res$, and returns the result (line 5).
If it is not the first time that Algorithm~\ref{Alg:Tuning} is called, then it retrieves the lower- ($l$) and upper- ($u$) bounds of threshold in $res$.
If the number of clusters in the current threshold is not equal to $u$, then it discovers a cluster for a threshold in between.
To avoid running this algorithm infinitely, we calculated the value in between by rounding the value by having two digits after the decimal points. 
It does the same for $l$, and it returns the result finally.
This algorithm tune the threshold parameter through a half-interval search.
\section{Implementation}\label{Sec:Implementation}

The approach is implemented and is available as a part of a python library, called \textit{processmining}. 
The source is available in Github~\footnote{\url{https://github.com/jalaliamin/processmining}}, and the library is available in PyPI - which enables users to install and use it easily by running the pip command~\footnote{\texttt{pip install processmining}}, if python and PM4Py are installed.
The library aims to provide more functionalities to perform process mining using python and other libraries like PM4Py. 
The codes to repeat the running example and evaluation can be found at Github~\footnote{\url{https://github.com/jalaliamin/ResearchCode/tree/main/ot-clustering-markov-dfm-ocpm}}.

\begin{figure}[t!] 
	\begin{minipage}{1\linewidth}
		\includegraphics[width=1\linewidth]{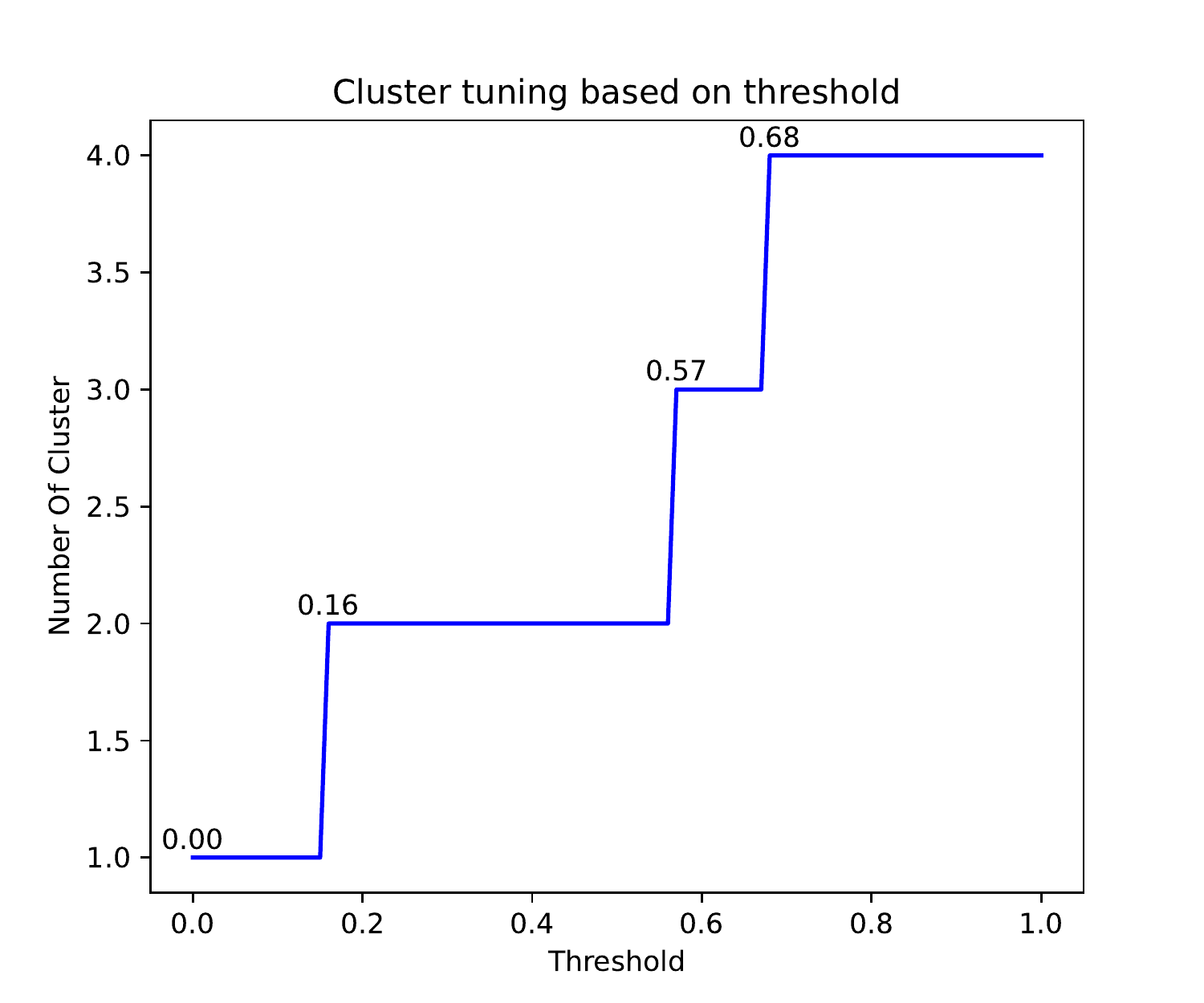}
	\end{minipage}%
	\caption
	{%
		The result of cluster tuning for \figurename~\ref{Fig:DFM} using implemented library.%
		\label{Fig:Table1_Tuning}%
	}%
\end{figure}

\figurename~\ref{Fig:Table1_Tuning} shows the result of cluster tuning for DFM in \figurename~\ref{Fig:DFM}, where it discovered four sets of clusters. 
In this figure, the x- and y- axes represent the threshold and number of retrieved clusters for the threshold parameter, respectively.

\begin{figure}[t!] 
	\begin{minipage}{1\linewidth}
		\begin{minipage}{.5\linewidth}
			\subfloat[0 threshold]
			{\includegraphics[width=1\linewidth]{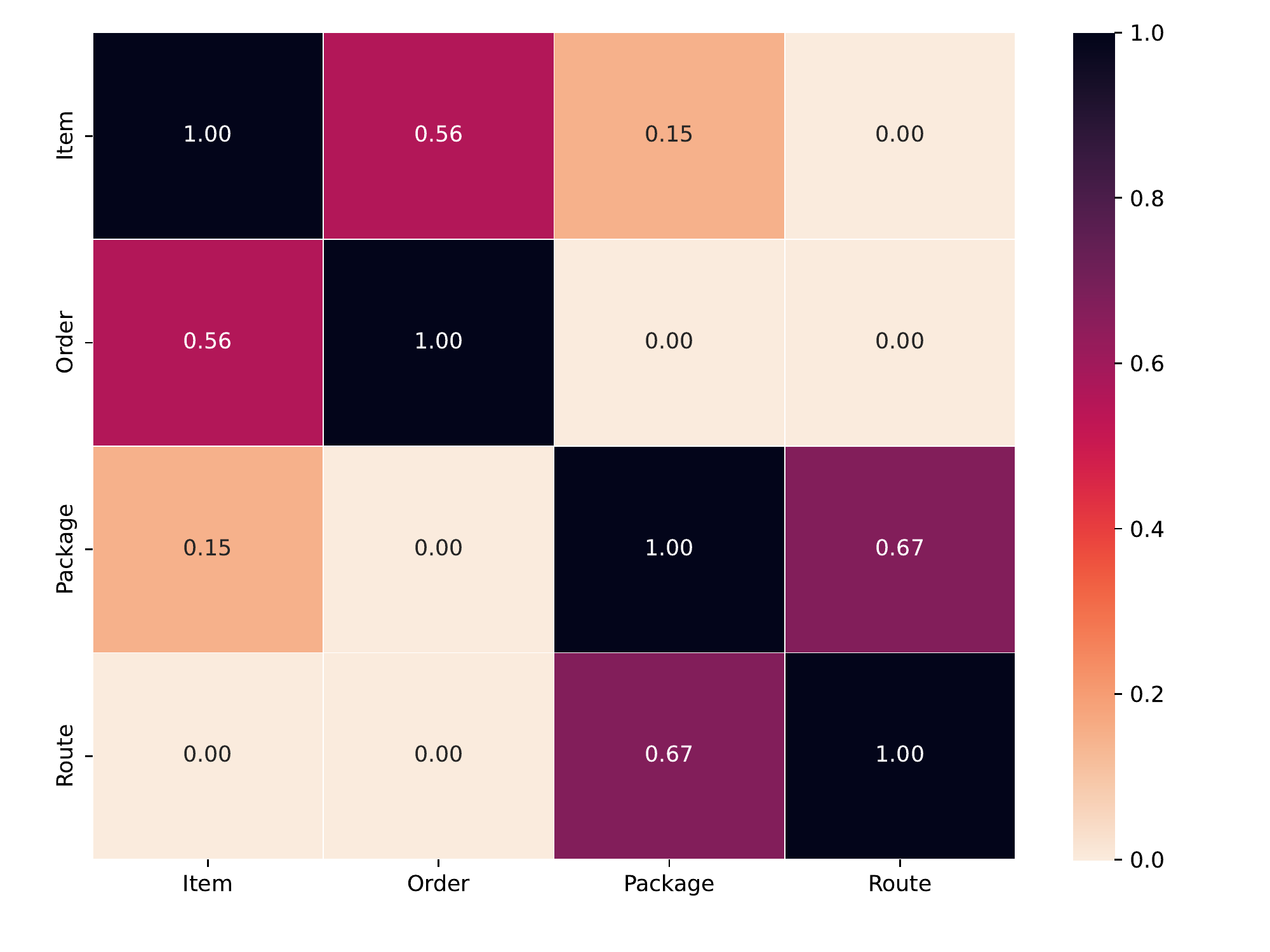}}\\
			\subfloat[0.16 threshold]
			{\includegraphics[width=1\linewidth]{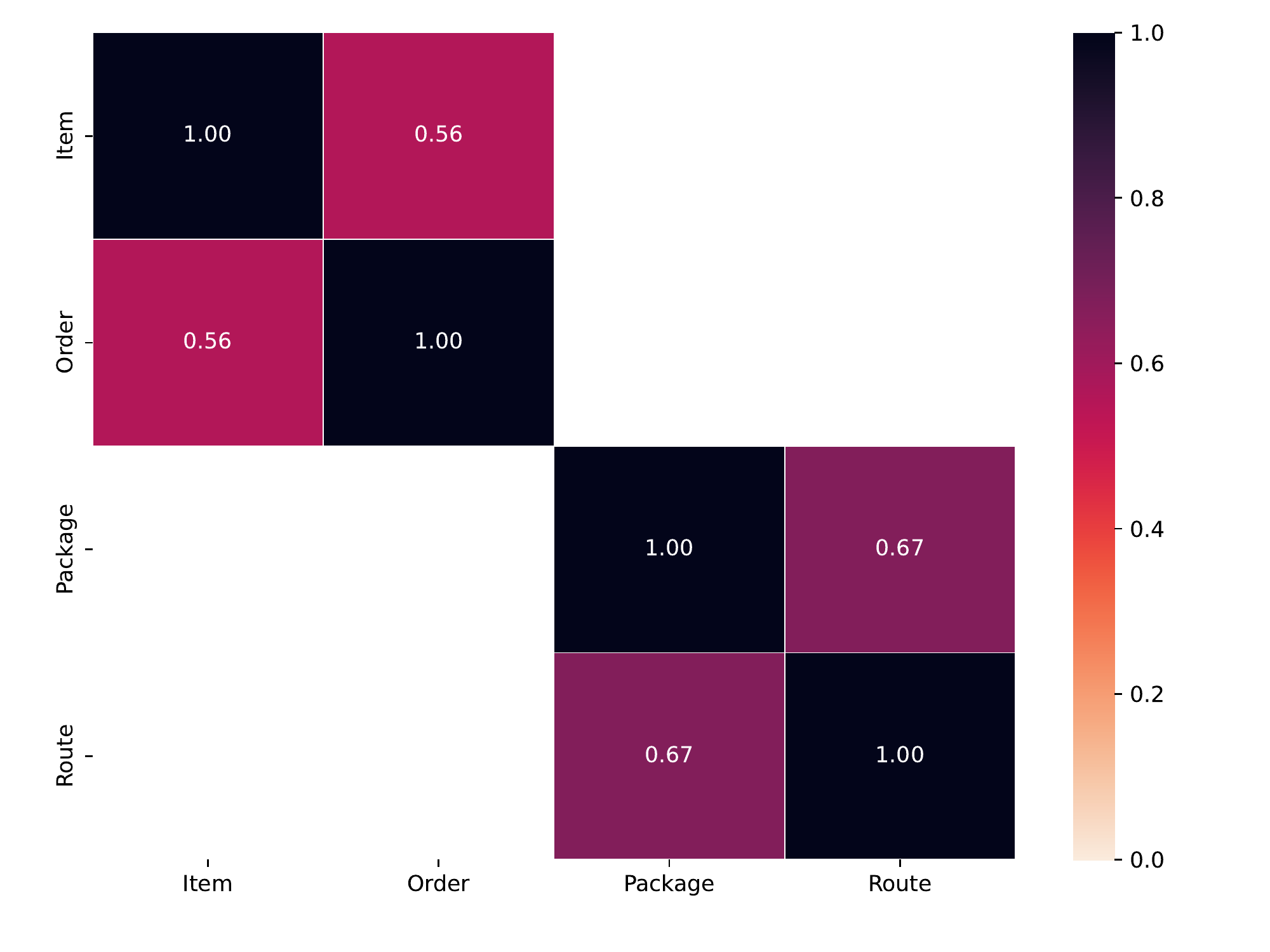}}
		\end{minipage}%
		\hfill
		\begin{minipage}{.5\linewidth}
			\subfloat[0.57 threshold]
			{\includegraphics[width=1\linewidth]{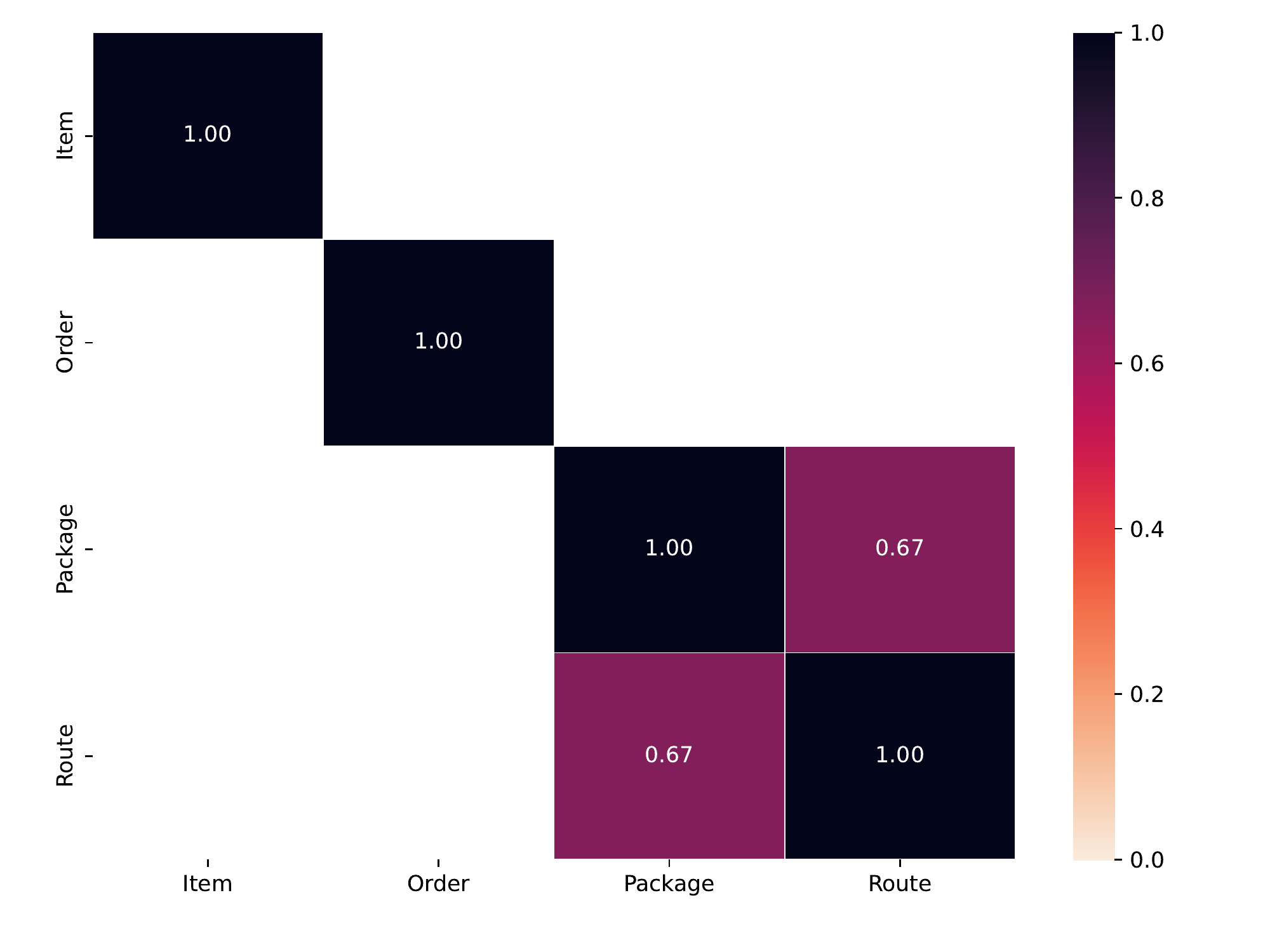}}%
			\\
			\subfloat[0.68 threshold]
			{\includegraphics[width=1\linewidth]{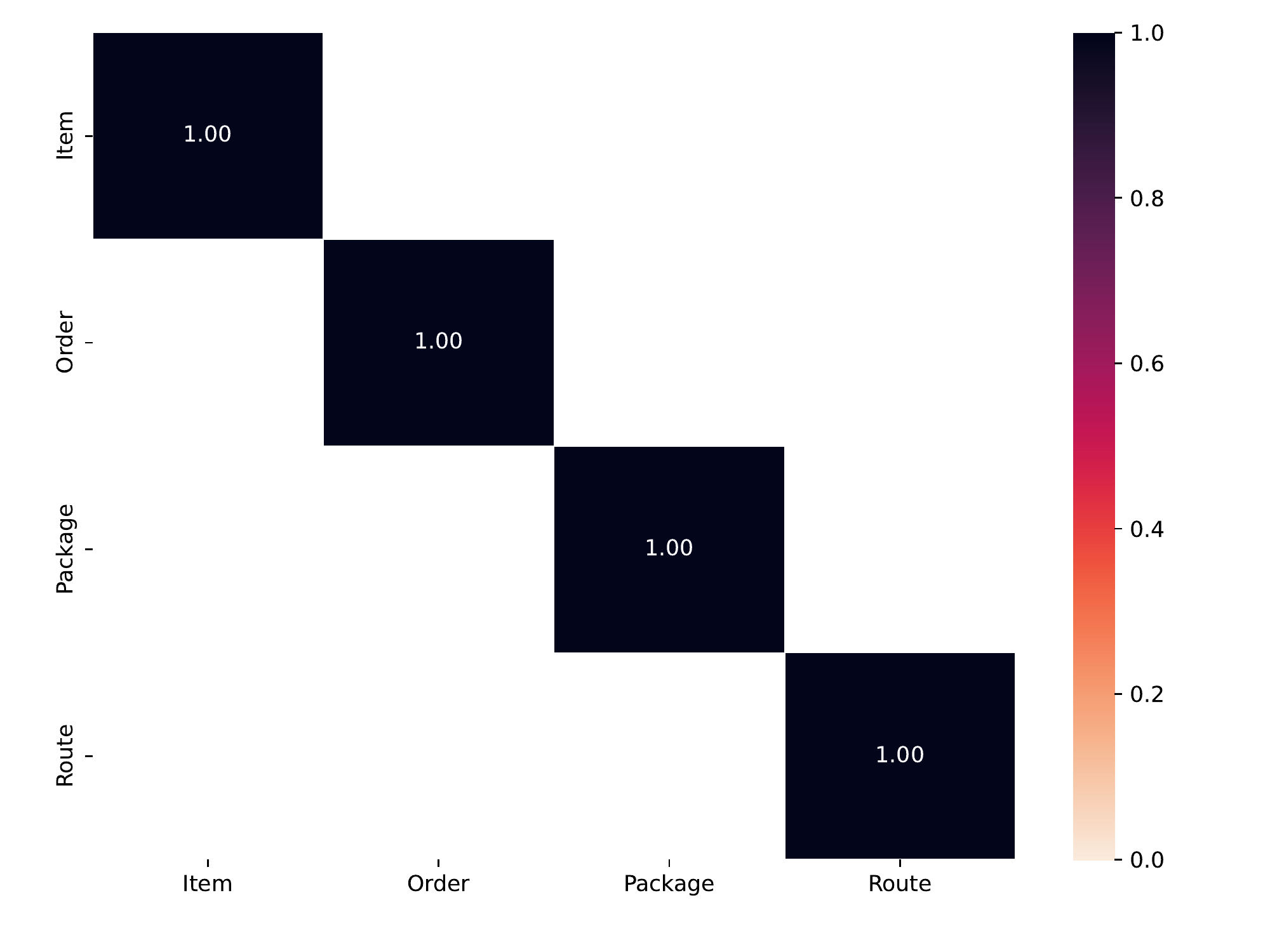}}%
		\end{minipage}%
	\end{minipage}%
	\caption
	{%
		The similarity matrices for identified clusters in \figurename~\ref{Fig:Table1_Tuning}.%
		\label{Fig:Table1_SimilarityMatrix}%
	}%
\end{figure}

\begin{figure}[t!] 
	\begin{minipage}{1\linewidth}
		\subfloat[DFM for the cluster that include Order and Item object types]
		{\includegraphics[width=1\linewidth]{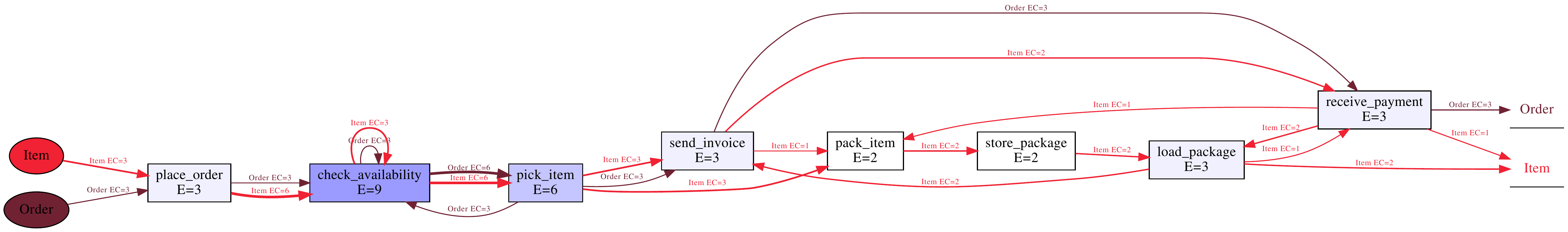}}
	\end{minipage}%
	\newline
	\begin{minipage}{1\linewidth}
		\subfloat[DFM for the cluster that include Package and Route object types]
		{\includegraphics[width=1\linewidth]{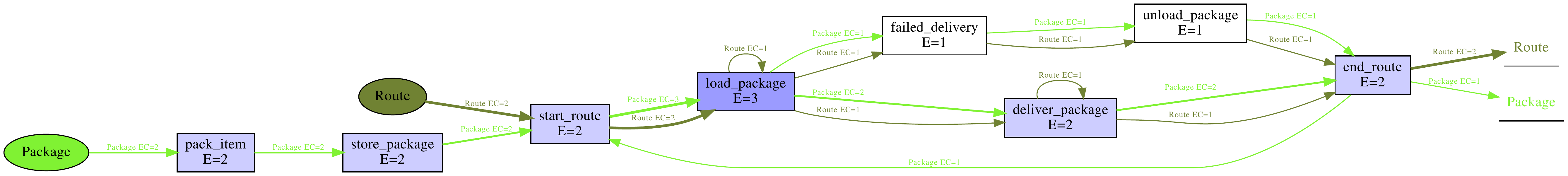}}
	\end{minipage}%
	\caption{Discovered DFMs based on two identified clusters by a similarity threshold of 0.16. The figure is made intentionally small just to show supporting the separation of similar object types.}
	\label{Fig:Table1_SeparatedProcessModels_0.16}
\end{figure}

The similarity matrix of these sets of clusters is plotted in \figurename~\ref{Fig:Table1_SimilarityMatrix}, where each sub-figure shows the similarity matrix for one threshold. 
As it can be seen from the sub-figures, the number of clusters in each set will be changed by changing the threshold. 
For example, if we set the threshold to 0.16, then it will return two clusters, i.e., $\{\{Item, Order\}, \{Package, Route\}\}$.
Flattening the OCEL based on these clusters can help discover process models with similar control behavior, as shown in \figurename~\ref{Fig:Table1_SeparatedProcessModels_0.16}.
This figure is made intentionally small only to show how the interconnected DFM in \figurename~\ref{Fig:DFM} will look in general when flattening the log based on similar object types. 
Such flattening still enables the study of the connection among similar object types, yet focusing on the related ones. 
The code for reproducing this experiment can be found in the Github~\footnote{\url{https://github.com/jalaliamin/ResearchCode/blob/main/ot-clustering-markov-dfm-ocpm/running-example.ipynb}}. 

\section{Evaluation}\label{Sec:Evaluation}

This section evaluates the presented approach using the given implementation on a Purchase to Pay (P2P) object-centric event log file.
For the evaluation, \textit{SAP ERP IDES instance - P2P log} file is used which is provided by \url{http://ocel-standard.org}~\cite{ghahfarokhi2021ocel}. 
This log file records the events for the Purchase to Pay process, and it contains 24,854	events and 9 object types.

These steps are followed to evaluate the approach. 
\textit{First}, sets of clusters are identified by applying this technique. 
\textit{Second}, some of the identified clusters are evaluated by flattening the log based on clustered object types. 
\textit{Third}, the log is flattened based on each object type, and a process model is discovered using the inductive miner for each flattened log.
For each pair of object types, their corresponding discovered models using inductive miner are compared using the footprint analysis technique.
\textit{Finally}, the result of the footprint analysis is compared with identified clusters. 

\subsection{Cluster discovery}

\begin{figure}[t!]
	\centering
	\begin{minipage}[b]{1\linewidth}
		\subfloat[threshold parameter tuning]
		{\includegraphics[width=1\linewidth]{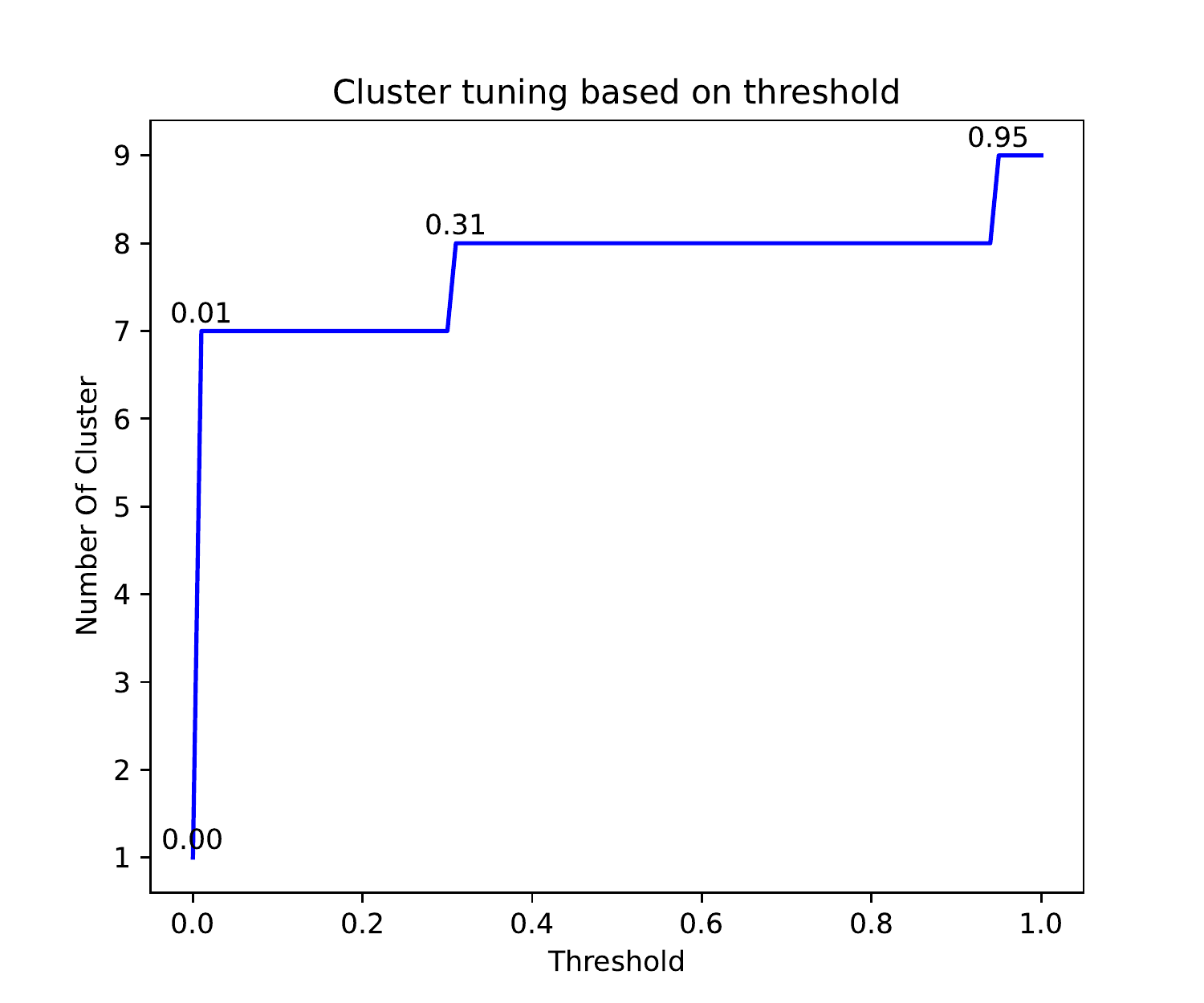}}
	\end{minipage}%
	\hfill
	\begin{minipage}[b]{1\linewidth}
		\subfloat[threshold=0.01]
		{\includegraphics[width=1\linewidth]{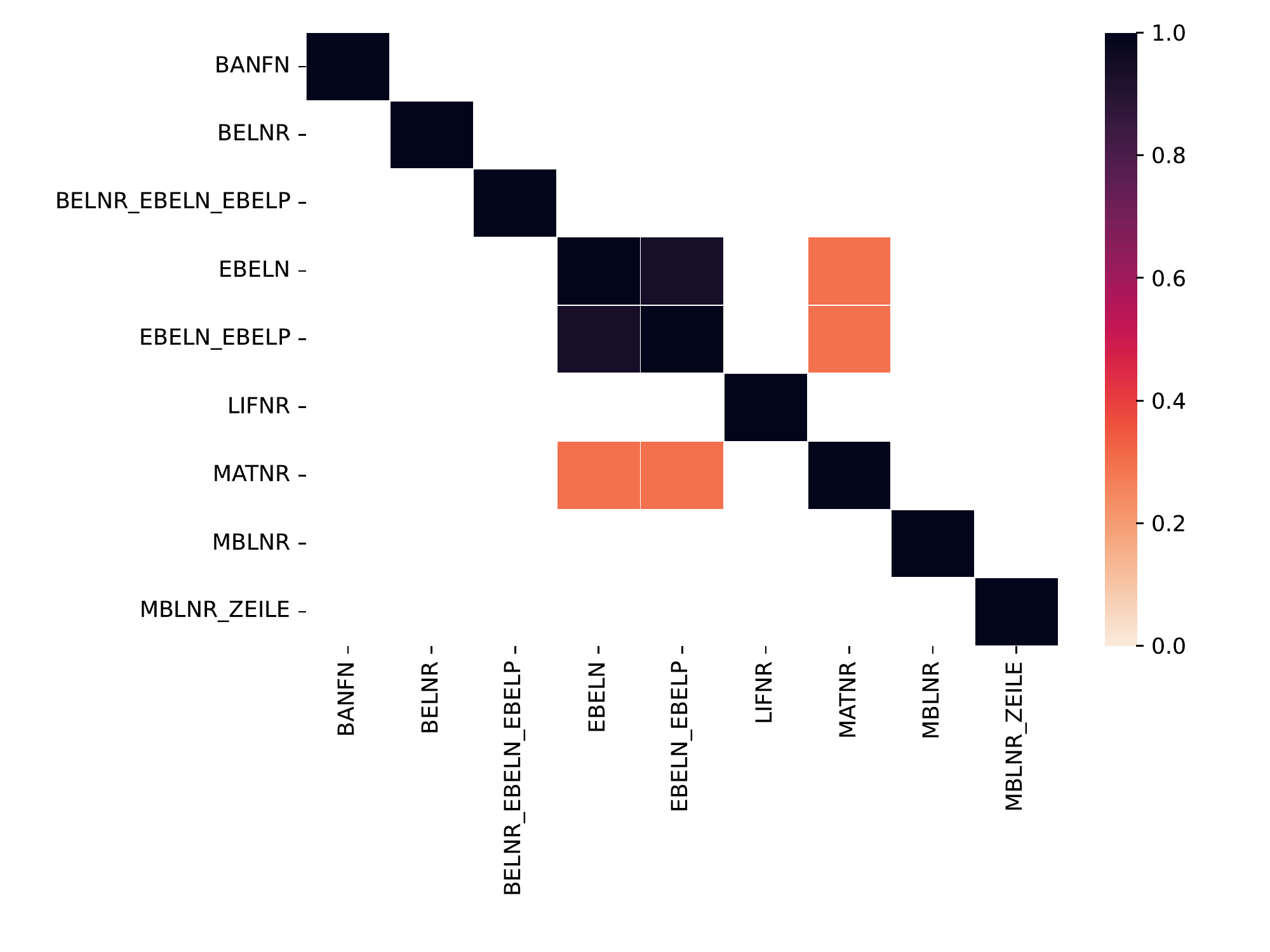}}
	\end{minipage}%
	\hfill
	\begin{minipage}[b]{1\linewidth}
		\subfloat[threshold=0.31]
		{\includegraphics[width=1\linewidth]{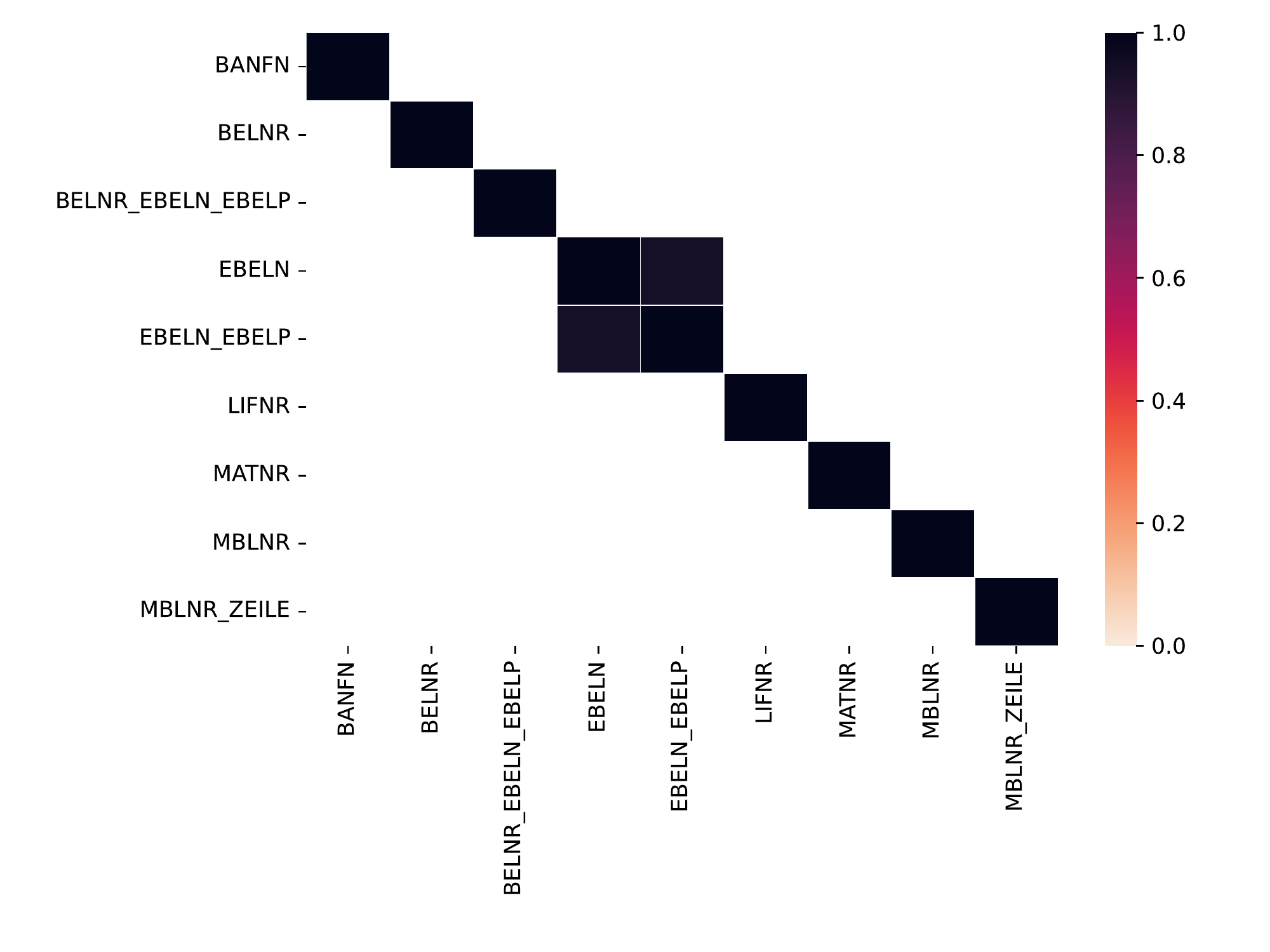}}\\
	\end{minipage}%
	\caption
	{%
		Cluster discovery result for p2p process.%
		\label{Fig:p2p_cluster_discovery}%
	}%
\end{figure}

This section presents the result of first and second steps in evaluating the proposed approach.
\figurename~\ref{Fig:p2p_cluster_discovery}(a) shows the result of threshold parameter tuning, where four different thresholds have been identified to discover different sets of clusters.
The first threshold, i.e., zero, will classify all object types into one cluster, and the last one will classify each object type in one cluster. 
Thus, we only present the two similarity matrices for the two remaining sets of clusters in \figurename~\ref{Fig:p2p_cluster_discovery}(b) and (c).

As can be seen in \figurename~\ref{Fig:p2p_cluster_discovery}(a), setting the threshold to 0.01 will result in 7 clusters. The similarity matrix in  \figurename~\ref{Fig:p2p_cluster_discovery}(b) shows that all object types except EBELN, EBELN\_EBELP, and MATNR are classified into their own clusters, meaning that they do not share any similar behavior. 

\begin{figure}[t!]
	\centering
	\begin{minipage}[b]{1\linewidth}
		\subfloat[Discovered DFM showing distinct behaviour of identified clusters when setting threshold to 0.01.]
		{\includegraphics[width=1\linewidth]{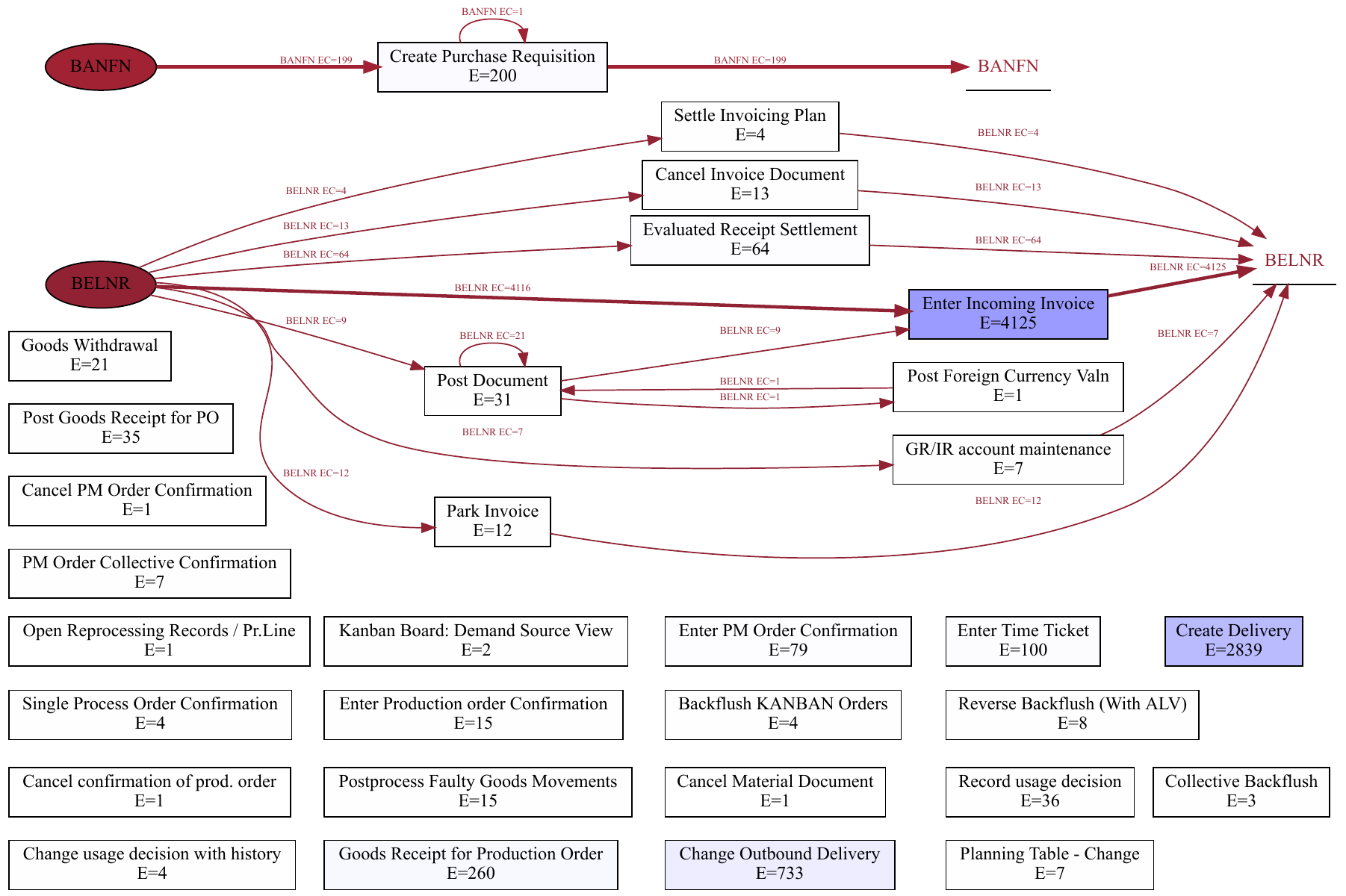}}
	\end{minipage}%
	\hfill
	\begin{minipage}[b]{1\linewidth}
		\subfloat[Discovered DFM showing similarity between behaviour for EBELN and  EBELN\_EBELP object types.]
		{\includegraphics[width=1\linewidth]{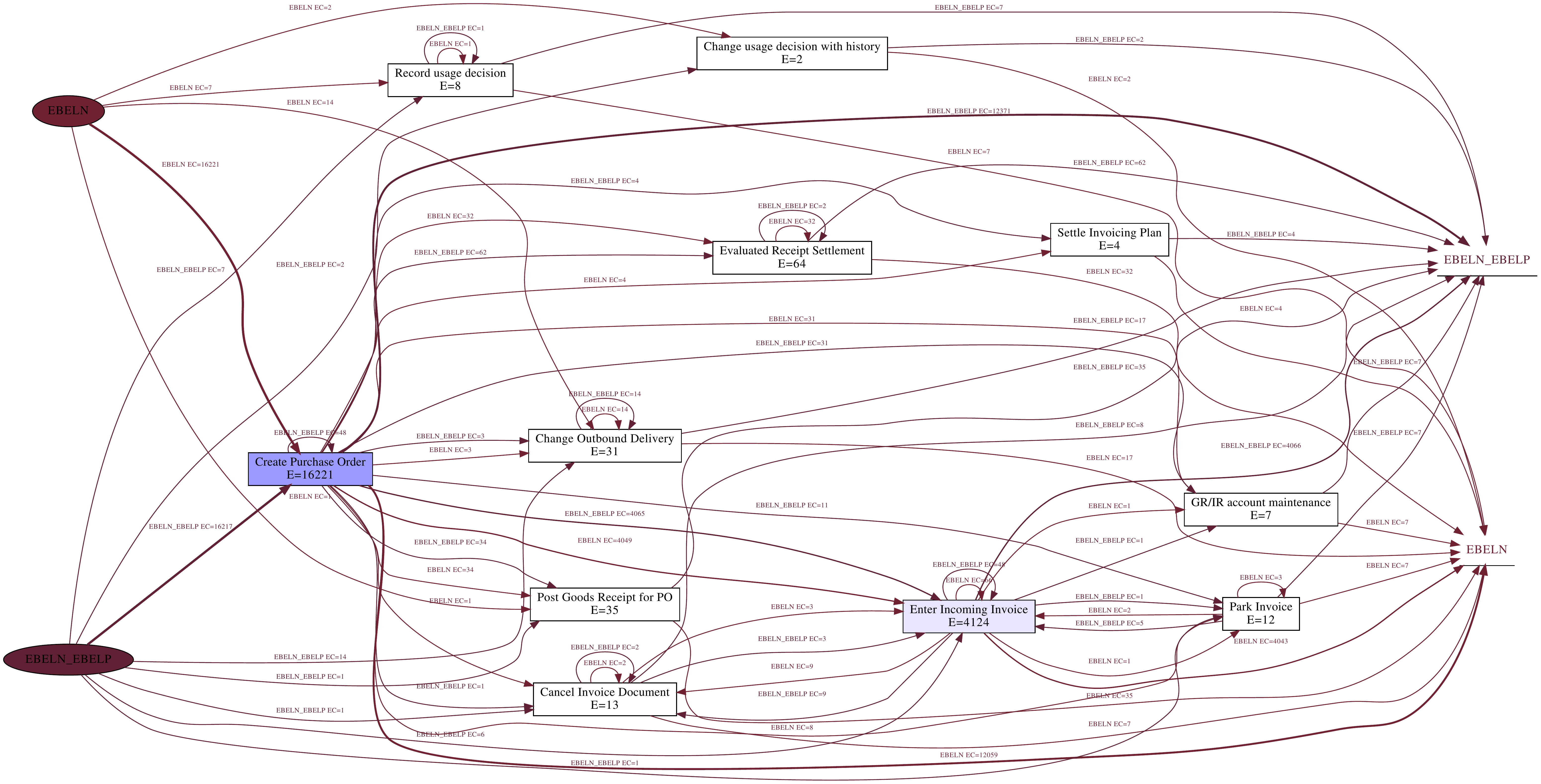}}
	\end{minipage}%
	\caption
	{%
		Discovered DFM for different object types. These processes are made intentionally small to show the validity of the result, and they are not meant to be read in detail.%
		\label{Fig:p2p_OnePercentThresholdClusters}%
	}%
\end{figure}

This finding can be validated by flattening the log based on these object types and discovering one process model, shown in \figurename~\ref{Fig:p2p_OnePercentThresholdClusters}(a).
The process is made intentionally small to show that there are unconnected tasks in addition to some disconnected control flow for different object types. 
The result confirms that these objects do not share similar behavior. 
Indeed, except for BANFN and BELNR object types, we do not see any occurrence of two consequent events for other object types.
The control flow for BANFN and BELNR object types also do not share any task, so they are distinct.

Increasing the threshold to 0.31 will discover a cluster with two object types, i.e.,  EBELN, EBELN\_EBELP. 
The process model, which is discovered by flattening the log based on these two object types, shows very similar behavior among them (see  \figurename~\ref{Fig:p2p_OnePercentThresholdClusters}(b)).

\subsection{Footprint analysis for flattened logs}

\begin{figure}[t!] 
	\begin{center}
		\includegraphics[width=1\linewidth]{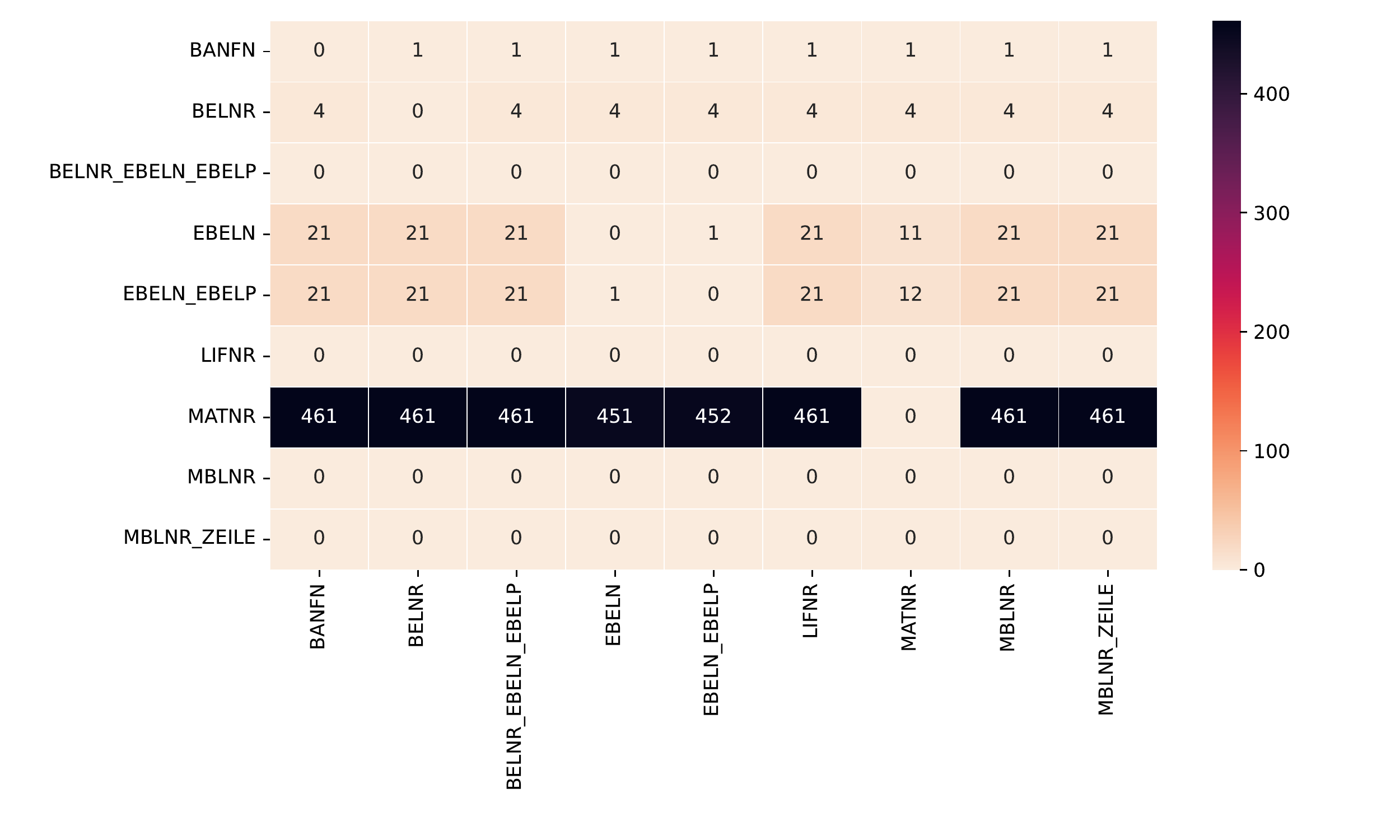}
		\caption{The calculated similarity between discovered process models using inductive miner by flattening the log based on each object type.}
		\label{Fig:p2p_Evaluation_footprint}
	\end{center}
	\vspace{-1\baselineskip}
\end{figure}

This section presents the result of the remaining steps in evaluating the proposed approach.
\figurename~\ref{Fig:p2p_Evaluation_footprint} shows the result of the conformance checking, where rows and columns represent object types, and cells represent the conformance of discovered process models using inductive miner by flattening the log - based on each object type.

As it can be seen, the highest difference in footprints belongs to MATNR, which has been identified as one cluster when setting the threshold to 0.31.
Taking this object type apart, the footprint difference for EBELN and EBELN\_EBELP has the highest difference with other object types while minimum difference to each other. 
This result aligns with the identification of the cluster that contains these two object types when setting the threshold between 0.31 and 0.94.
The code for this experiment is available in Github~\footnote{\url{https://github.com/jalaliamin/ResearchCode/blob/main/ot-clustering-markov-dfm-ocpm/p2p.ipynb}}.

\section{Conclusion}\label{Sec:Conclusion}

This paper introduced a new approach to cluster similar case notions by defining Markov Directly-Follow Multigraph.
The graph is used to define an algorithm for discovering clusters of similar case notions based on a threshold. 
The paper also defined a threshold tuning algorithm to identify sets of different clusters that can be discovered based on different levels of similarity. 
Thus, the cluster discovery does not merely rely on analysts' assumptions. 
The approach is implemented and released as a part of a python library, called \textit{processmining}, and it is evaluated through a Purchase to Pay (P2P) object-centric event log file. 
Some discovered clusters are evaluated by discovering Directly Follow-Multigraph by flattening the log based on the clusters. 
The similarity between identified clusters is also evaluated by calculating the similarity between the behavior of the process models discovered for each case notion using inductive miner based on footprints conformance checking.

This approach can be used to define an object-centric process discovery algorithm that takes the similarity of object types into account when discovering process models from object-centric event logs, which will be a future direction of this work. 

\bibliographystyle{plain}
\bibliography{References}

\end{document}